\documentclass[journal,twoside,web]{ieeecolor}
\usepackage{generic}
\usepackage{cite}
\usepackage{amsmath,amssymb,amsfonts}
\usepackage{algorithmic}
\usepackage{graphicx}
\usepackage{textcomp}
\usepackage{tabularx}      
\usepackage{booktabs}
\usepackage{multirow}
\usepackage[caption=false,font=normalsize,labelfont=sf,textfont=sf]{subfig}
\def\BibTeX{{\rm B\kern-.05em{\sc i\kern-.025em b}\kern-.08em
    T\kern-.1667em\lower.7ex\hbox{E}\kern-.125emX}}
\begin{document}
\title{Exact Fit Attention in Node-Holistic Graph Convolutional Network for Improved EEG-Based Driver Fatigue Detection}
\author{ Meiyan Xu$^{1}$, Qingqing Chen$^{1}$, Duo Chen$^{*}$,   Yi Ding, Jingyuan Wang, Peipei Gu, Yijie Pan, Deshuang Huang, \IEEEmembership{Fellow, IEEE} and Jiayang Guo$^{*}$
\thanks{This work was supported by the National Natural Science Funds of China (62006100), the Fujian Provincial Natural Science Foundation of China (2024J01009), the Fundamental Research Funds for the Chinese Central Universities of China (0070ZK1096), the Young Tech Innovation Leading Talent Program of Ningbo City under Grant (2023QL008), the Fujian Provincial Natural Science Foundation of China (2023J01921),  \\
(*Corresponding authors: Duo Chen; Jiayang Guo.)}
\thanks{Meiyan Xu and Qingqing Chen are with the Department of Computer Science and the Institute of Human Factors Engineering, Minnan Normal University, Zhangzhou 363000, China.}
\thanks{Duo Chen is with the School of Artificial Intelligence and Information Technology, Nanjing University of Chinese Medicine, Nanjing 210023, China (e-mail: 380013@njucm.edu.cn).}
\thanks{Yi Ding is with the College of Computing and Data Science, Nanyang Technological University, Singapore.}
\thanks{Jingyuan Wang is with the School of Digital Media and Design Arts, Beijing University of Posts and Telecommunications, Beijing 100088, China.}
\thanks{Peipei Gu is with the College of Software Engineering, Zhengzhou University of Light Industry, Zhengzhou 450000, China.}
\thanks{Yijie Pan is with the Department of Computer Science and Technology, Tsinghua University, Beijing 100084, China, and also with the Eastern Institute for Advanced Study, Eastern Institute of Technology, Ningbo 315200, China.}
\thanks{Deshuang Huang is with Ningbo Institute of Digital Twin, Eastern Institute of Technology, Ningbo 315200, China.}
\thanks{Jiayang Guo is with the National Institute for Data Science in Health and Medicine, Xiamen University, Xiamen 361005, China (e-mail: guojy@xmu.edu.cn).}
}

\maketitle

\begin{abstract}
EEG-based fatigue monitoring can effectively reduce the incidence of related traffic accidents. In the past decade, with the advancement of deep learning, convolutional neural networks (CNN) have been increasingly used for EEG signal processing.
However, due to the data's non-Euclidean characteristics, existing CNNs may lose important spatial information from EEG, specifically channel correlation. 
Thus, we propose the node-holistic graph convolutional network (NHGNet), a model that uses graphic convolution to dynamically learn each channel's features. With exact fit attention optimization, the network captures inter-channel correlations through a trainable adjacency matrix. The interpretability is enhanced by revealing critical areas of brain activity and their interrelations in various mental states. 
In validations on two public datasets, NHGNet outperforms the SOTAs. Specifically, in the intra-subject, NHGNet improved detection accuracy by at least 2.34\% and 3.42\%, and in the inter-subjects, it improved by at least 2.09\% and 15.06\%.
Visualization research on the model revealed that the central parietal area plays an important role in detecting fatigue levels, whereas the frontal and temporal lobes are essential for maintaining vigilance.
\end{abstract}


\section{Introduction}
Of all road fatalities, 10-30\% are caused by fatigue \cite{jaydarifard2023driver}. A significant emphasis within automotive safety engineering is placed on the development of efficient driver fatigue monitoring systems to mitigate the risk of accidents\cite{zhu2021vehicle,zeng2020instanceeasytl}. 
Studies aiming to detect driver weariness increasingly use scalp electroencephalogram (EEG), making that affordable with high temporal resolution \cite{cui2022eeg}. EEG data is often represented in 2D time series, images, and graph formats. In 2D time series representation, neural network input layers commonly include temporal, spatial, and spatial-temporal mixed convolutional layers. Researchers extract spatial-temporal characteristics from specific areas using this method \cite{wang2023identifying,lee2023lstm}, but they miss channel spatial distribution. The second representation method involves using images and arranging channels in a 2D framework according to their relative positions on the surface of the brain, thereby highlighting the spatial positions relative to each other \cite{gao2023sft,khan2023novel}. Both representation methods presuppose that EEG data reside in Euclidean space, a premise that does not align with reality.

In recent years, there has been an increasing trend towards representing EEG data graphically \cite{hou2022gcns,ho2023self}. In this approach, EEG signals are conceptualized as a graph, where channels serve as nodes and their spatial distances or correlations form the edges. This method is particularly suitable for analyzing interactions between different brain regions, as it directly focuses on the relationships between channels.

Some researchers \cite{ko2020vignet,geirnaert2020fast} manually extract characteristics and apply machine learning for classification, but this needs previous knowledge and may overlook important information \cite{luo2023dual}. In contrast, end-to-end deep learning approaches can learn features directly from raw data, thus more effectively capturing the intricate patterns in the data \cite{li2024cwstr}. These approaches aim to extract temporal and spatial information from raw data \cite{ding2023lggnet}. Previous 1-dimensional (1D) convolutional neural networks (CNNs) \cite{lawhern2018eegnet,wang2023eeg} and multi-scale 1D CNNs \cite{tang2023motor,wei2023ms} commonly excel in temporal analysis. In contrast, some methods focus on spatial analysis. For example, 1D CNNs \cite{liu2022fbmsnet,ding2020tsception} and 2-dimensional (2D) CNNs \cite{jiao2018deep,jung2019utilizing,liu2023eeg} were used to extract local spatial information achieving promising results in emotion detection.
However, these methods also have some drawbacks, particularly ignoring spatial distribution information between channels. So, researchers have begun to focus on the spatial relations of channels by expressing EEG data in 3-dimensional (3D) \cite{zhao2019multi,jiao2023domain,wang2024braingridnet} and using 3D CNNs for learning in recent years \cite{huang2023spatio,park2023spatio}.

The methods mentioned above based on CNNs excel in handling structured, grid-based data, their ability to capture spatial correlations often relies on predefined or fixed neighborhood structures, which to some extent limits their flexibility. In contrast, graph convolutional networks (GCN) \cite{jia2023end,cai2023brain,zhang2024novel} demonstrate greater flexibility and adaptability when processing EEG data in non-Euclidean spaces. Numerous GCN models have been proposed and studied for EEG decoding \cite{wang2020linking,ding2023lggnet}. However, current graph-convolution-based EEG decoding models face challenges. During the process of transforming data or features into graph representations, spatial information becomes mixed and intertwined~\cite{WANG2024102229}. In addition, some models tend to focus only on capturing the spatial features of specific channel groups rather than conducting a deep analysis and extraction of information from each EEG channel, followed by a global learning process~\cite{TANG2024121915,9857562}. 

Hence, this paper introduces a node-holistic graph convolutional network (NHGNet), which considers the feature vector obtained from each EEG channel after a three-branch temporal feature extractor and exact fit attention (EF-attention) as a node in a graph network. Following that, it uses GCN to learn information for each channel and capture correlation with a global dynamic similarity adjacency matrix (GDSAM). Furthermore, we study brain area activity and correlations in various mental states.
This work makes the following key contributions:
\begin{itemize}
\item We introduce EF-attention, which captures the subtle differences between temporal points and channels, enhances the sensitivity of the model to features and provides a broader adaptive adjustment space than traditional attention mechanisms.
\item We present NHGNet, which captures spatial properties inside brain functional regions and uses trainable fully linked matrices to portray complicated connections between these areas.
\item We reveal alterations in brain area activity patterns and their interconnections among channels in distinct cognitive states.
\end{itemize}

\section{Related Works}

GCN demonstrates exceptional performance in several aspects of the EEG. Jang et al. used EEG graph structures to classify videos by studying channel location and correlation \cite{jang2018eeg}. For emotion recognition, Song et al. suggested a dynamic GCN that uses trainable adjacency matrices to represent EEG channel functional correlations\cite{song2021variational}. STS-HGCN \cite{li2021spatio} extracts the spatio-temporal feature of epileptic seizures, showcasing the medical potential of GCN. Zhang et al. have created a sparse DGCNN model with graph constraints to localize and sparsify weights, improving performance \cite{zhang2021sparsedgcnn}. AMCNN-DGCN \cite{wang2020linking} and LGGNet \cite{ding2023lggnet} have effectively employed GCN to achieve significant advancements in the field of fatigue driving.

However, the aforementioned models exhibit certain limitations in capturing spatial features, primarily manifested in two areas: Firstly, the feature extraction process before transforming data into graphical form often results in ambiguous information representation within the graphical nodes. Second, there is a tendency for these models to focus on specific combinations of channels while overlooking the significance of conducting detailed analysis and comprehensive information extraction on individual EEG channels. Such issues could diminish the effectiveness of spatial feature learning by the models.

\section{Proposed Model}
\begin{figure*}[t]
\begin{center}
\centerline{\includegraphics[width=\textwidth]{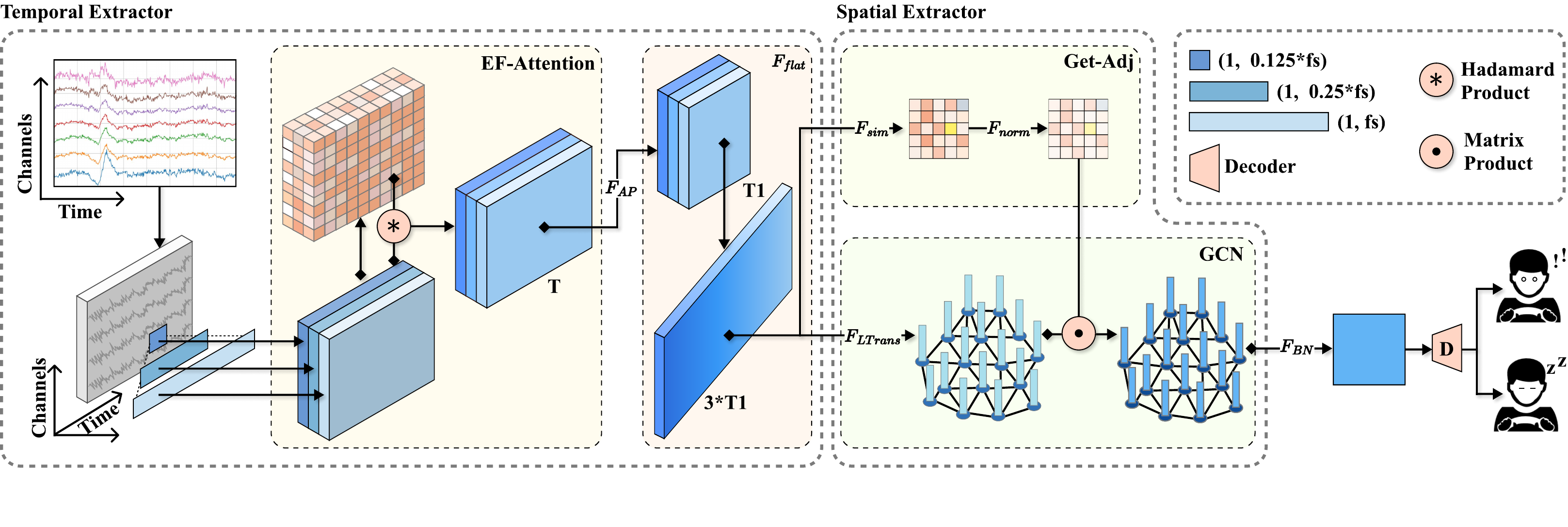}}
\caption{Network structure of NHGNet. fs denotes the ultimate sampling rate. $\text{T}_1$ represents the new data dimension produced by dimensionality reduction. BN stands for batch normalization.}
\label{fig:model}
\end{center}
\end{figure*}

This section discusses the NHGNet structure, as shown in Fig. \ref{fig:model}. Section \ref{subsec:temporal_extractor} explores the temporal extractor; Section \ref{subsec:spatial_extractor} describes the spatial extractor and classifier; and Section \ref{subsec:training_methodology} elaborates on the specific training strategies in intra-subject and inter-subject.

\subsection{Temporal Extractor}
\label{subsec:temporal_extractor}
This session describes a temporal feature extractor that includes three main components: multiscale temporal depthwise (MTD) convolution, EF-attention, and a logarithm energy (LogEnergy) layer.

\textbf{MTD convolution:} The MTD convolution is designed to extract characteristics from various frequency ranges using deepwise convolution layers with different convolution kernel sizes due to the significant correlation between frequency band and fatigue level. Deepwise CNNs can separately process each channel, allowing for the extraction of distinct information from each EEG channel.

As demonstrated in the Fig. \ref{fig:model}, let $X$ be an EEG signal represented by a vector $X = [x_1, x_2, x_3, \ldots, x_N] \in \mathbb{R}^{N \times T}$, where $N$ is the number of EEG channels and $T$ is the number of sample points per channel. The EEG is first processed using a three-branch deepwise convolutional network, with each branch utilizing convolutional kernels of various sizes. In this scenario, the output of each branch is \(d_i = F_i(X)\), where \(F_i(\cdot)\) is the $i$-th deepwise convolution. The convolutional kernel size for \(F_i(\cdot)\) is specified as \((1, 0.5^i \cdot fs)\), with \(i \in \{1, 2, 3\}\). Here, \(fs\) represents the sampling rate of the data. By doing this, we can combine features from different time scales to create a composite feature map $D = C(d_i)$, where $C(\cdot)$ represents the concatenation of tensors.

\textbf{EF-attention:} The EF-attention mechanism initially computes channel weights through pointwise convolution, followed by the application of the $\text{Tanh}$ activation function to modify and adjust the weights for each channel. The pointwise convolution employs the smallest possible convolution kernel and stride to meticulously capture the subtle variations across time points and channels. Subsequent weight adjustments are made using the $\text{Tanh}$ activation function, which notably benefits from its scaling range of $[-1, 1]$. Compared to more commonly used activation functions in attention mechanisms, such as $\text{Softmax}$ and $\text{Sigmoid}$, the $\text{Tanh}$ activation function offers a broader adaptive adjustment space. This feature allows the EF-attention mechanism to modulate channel features more finely and effectively, thereby optimizing feature representation within the network.

Actually, $W_A = \tanh(\text{pointwise} (D))$ defines this step. Here, \( W_A \) is the weight map produced by the \( \tanh \) activation function, whose output falls between $[-1, 1]$. Subsequently, a Hadamard product merges the weight map $W_A$ with $D$ to build a weighted feature map $H\in \mathbb{R}^{N \times T \times 3}$, where $H = D \circ W_A$.

\textbf{LogEnergy Layer:} We provide an experimental layer named ``LogEnergy Layer'' that is intended to improve EEG processing. This component conducts a series of extensive processing tasks, such as batch normalization (BN), activation, and the calculation of representations through logarithmic and squaring operations, which significantly improves the feature representation of input data. This joint equation can be used to represent this process:
\begin{equation*}
P = \log_{10}\left(\text{AvgPool}\left(\left(\text{LeakyReLU}\left(\text{BN}(H)\right)\right)^2\right)\right).
\end{equation*}
BN improves training stability and accelerates model convergence. The LeakyReLU activation function introduces nonlinearity, which keeps the gradient from dissipating, and then calculates the logarithm of the average pooled value after squaring. The kernel size of the AvgPool is set to half the sampling rate with a stride size of 1, which not only improves feature representation but also downscales the data to transform temporal features into a more compact and information-rich format.

\subsection{Spatial Extractor}
\label{subsec:spatial_extractor}
This section will shift its attention from temporal extractors to spatial extractors and classifiers. The spatial extractor comprises a GCN and a flexible channel correlation (FCC) unit.

We flatten the time feature vector $P$ across $N$ EEG channels, considering each channel as a node in the EEG graph. To clarify, we convert $P$ into $G_{\text{raw}}$ using the function $\phi_{\text{reshape}}(\cdot)$, which can be expressed as $G_{\text{raw}} = \phi_{\text{reshape}}(P) = [G_1, G_2, G_3, \ldots, G_N]^T$. Here, \(G_j\) (where \(j \in \{1, 2, \ldots, N\}\)) denotes the dynamic temporal representation acquired at each channel. According to this, we will conduct detailed learning on each channel individually and holistic dynamic learning on $G_{\text{raw}}$.

\textbf{GCN:} This approach is intended to capture the detailed properties of each channel in EEG data. In our network, the GCN receives input features of a specific size and maps them to the required output feature size. The size of the input features corresponds to the temporal feature dimension of the EEG signal, whereas the size of the output features $F$ is dictated by the desired level of feature abstraction ($F$ = 32 in this study).

By applying the GCN layer, we have effectively captured the dynamic features of each channel in the EEG signal as they vary over time. Let $w_{\text{detail}} \in \mathbb{R}^{N \times F}$ be the weight matrix responsible for transforming the input features into more advanced feature representations. Similarly, $b_{\text{detail}} \in \mathbb{R}^{1 \times F}$ is the bias term, initially set to zero but adjusted during the training process to optimize feature learning.

The detail feature map $G_{\text{detail}} \in \mathbb{R}^{N \times F}$ can be represented by the following graph convolution operation formula:
\begin{align*}
G_{\text{detail}} &= G_{\text{raw}}w_{\text{detail}} + b_{\text{detail}} \\
&= [G_1 w_{\text{detail}} + b_{\text{detail}}, \, \ldots, \, G_N w_{\text{detail}} + b_{\text{detail}}]^T \\
&= [G_{\text{detail}}^1, G_{\text{detail}}^2, G_{\text{detail}}^3, \ldots, G_{\text{detail}}^N]^T,
\end{align*}
where $G_{\text{detail}}^j$ represents the transformed feature representation of the $j$-th EEG channel after processing through the GCN.

The primary benefit of this method lies in its focus on the unique properties of each channel in the EEG data, with our network employing a carefully constructed weight matrix $w_{\text{detail}}$ to learn attributes of each channel and transform them into more expressive forms. This stage is essential because it allows us to understand how each channel individually contributes to overall brain activity, laying the groundwork for subsequent research with global feature learning.

\textbf{FCC unit:} The FCC unit creates a GDSAM $S \in \mathbb{R}^{N \times N}$ that depicts the interrelationships between channels in EEG data. This GDSAM has trainable characteristics, meaning it can adjust based on backpropagation of the loss function during the training process, thus adapting to specific classification tasks.

Initially, obtaining a preliminary adjacency matrix (AM) is required. Predicting the precise correlations between EEG channels is a substantial difficulty. Some studies base their approach on the physical distance between channels \cite{zhong2020eeg,du2022multi}, while others use phase locking values to construct the AM \cite{hsieh2022graph,shen2022multiscale}. However, these approaches frequently record the associations between channels from a single angle, ignoring other potential connections. Inspired by LGGNet \cite{ding2023lggnet}, we choose to describe the relationships between distinct EEG channels using the dot product. Given that the relationship between any two channels is reciprocal, we suppose that the global adjacency matrix (GAM) is undirected. Therefore, the global similarity adjacency matrix (GSAM) $S_{\text{base}} \in \mathbb{R}^{N \times N}$ is defined as a symmetric matrix, calculated as follows:
\begin{align*}
S_{\text{base}} &= G_{\text{detail}}\ G_{\text{detail}}^T \\
&= \begin{bmatrix} G_{\text{detail}}^1 \cdot G_{\text{detail}}^1 & \cdots & G_{\text{detail}}^1 \cdot G_{\text{detail}}^N \\ \vdots & \ddots & \vdots \\ G_{\text{detail}}^N \cdot G_{\text{detail}}^1 & \cdots & G_{\text{detail}}^N \cdot G_{\text{detail}}^N \end{bmatrix}.
\end{align*}
Due to the complexity of the brain functional network, in the FCC unit, we use a trainable mask board $M_{\text{base}} \in \mathbb{R}^{N \times N}$ to emphasize key connections in the GSAM $S_{\text{base}}$. This mask board is initialised using the Xavier uniform initialization method, which helps maintain gradient and signal stability across different layers of the neural network. Given that the GAM is undirected, the mask board $M$ we designed is also symmetric. Therefore, the final mask board $M$ used to capture channel correlations is defined as $M = M_{\text{base}} + M_{\text{base}}^T$. To ensure that each node retains its own feature information and the non-negativity of the GDSAM, we add self-loops and use the rectified linear unit (ReLU) activation function. Thus, the GDSAM $S_{\text{overall}}\in \mathbb{R}^{N \times N}$ can be calculated as follows:
$ S_{\text{overall}} = \text{ReLU}(S_{\text{base}} \circ M) + I $
where $I$ represents the identity matrix, used for adding self-loops. The GDSAM $S_{\text{overall}}$ is then normalized to balance the influence of different nodes in the graph. Let \(D\) be the degree matrix, whose diagonal elements are the sum of each row in the AM, i.e., \(D = \text{diag}\left(\sum_{q} S_{\text{overall}}^{pq}\right)\), where \(p, q \in \{1, 2, \ldots, N\}\), and the function \(\text{diag}(\cdot)\) is used to convert the summation elements into a diagonal matrix.
The final GDSAM \(S\) can be expressed as
$$S = D^{-\frac{1}{2}}\ S_{\text{overall}}\ D^{-\frac{1}{2}}.$$

The final comprehensive global feature is $G_{\text{overall}} = SG_{\text{detail}}$. The GDSAM $S$ reveals the interrelationships between different channels in the EEG data. $G_{\text{overall}}$ obtains more information about channel interactions, capturing the intricate interplay between them and offering a more comprehensive view of brain activity.

\textbf{Classifier}: After the spatial feature extractor completes its process, the output undergoes BN. Subsequently, its flattened representation is input into a linear layer, which generates the final classification output, denoted as $\hat{y}$.

The output of NHGNet uses cross-entropy loss $l$ to assess the discrepancy between true labels $y$ and predicted labels $\hat{y}$.
The loss function is defined as follows:
\begin{equation*}
l = -\sum_{i=1}^{2} y_i \cdot \log(\hat{y}_i) + \lambda_1\sum |\theta| + \lambda_2\sum \theta^2,
\end{equation*}
where $\theta$ represents all model parameters during training, while $\lambda_1$ and $\lambda_2$ are the L1 and L2 regularisation coefficients, respectively, used to prevent overfitting.

\subsection{Training Methodology}
\label{subsec:training_methodology}
Nested cross-validation is an excellent way to reduce bias in model evaluation. We use two layers of cross-validation: an outer loop and an inner loop. The outside loop utilizes 10-fold cross-validation to steer the experimental path, while the inner loop uses 3-fold cross-validation for optimum model selection. To improve experimental efficiency, an early stopping mechanism is utilized, specifically halting the experiment prematurely if the accuracy of the validation set does not increase after 20 consecutive epochs. Within the three rounds of cross-validation in the inner loop, the model with the best accuracy on the validation set is chosen and kept as the candidate model.

In intra-subject experiments, the candidate model is used directly as the final test model. In contrast, in inter-subject experiments, this model serves as a foundational model for later refinement via transfer learning. The transfer learning technique entails separating the target subject data into two parts: one half is used for fine-tuning throughout the transfer learning process, while the other half is set aside as test data. During this phase, the learning rate is reduced to one-tenth of its initial value. The transfer learning procedure ends after 50 training epochs or when the training accuracy reaches 100\%.

\section{Experimental Results}

\begin{table*}[h]
\caption{The experimental results of different models across two datasets in intra-subject comparisons. These results are expressed as the mean ± standard deviation, with the highest values highlighted in bold font. $p$-value of the improvement of NHGNet over the model: * indicating $p<0.05$, ** indicating $p<0.001$}
\label{tab:intra-subject}
\begin{center}
\begin{small}
\begin{tabularx}{\textwidth}{@{}@{}c c c X X X X X@{}}
\toprule
Datasets & Model & Year & Accuracy(\%) & Precision(\%) & Sensitivity(\%) & Specificity(\%) & F1-score(\%) \\
\midrule
\multirow{6}{*}{Dataset 1} & EEGNet & 2018 & 85.16$\pm$13.23 & 83.34$\pm$20.16 & 79.20$\pm$26.11 & 83.27$\pm$27.84 & 79.33$\pm$22.48 \\
& ICNN & 2022 & 86.61$\pm$7.78** & 91.13$\pm$9.18** & 80.70$\pm$16.14** & 87.65$\pm$19.36** & 84.12$\pm$9.63** \\
& ECNN & 2023 & 90.32$\pm$6.51** & 86.18$\pm$28.19** & 78.25$\pm$28.36** & 89.55$\pm$19.98** & 80.99$\pm$26.74** \\
& LGGNet & 2023 & 90.76** & \qquad\,\ - & \qquad\,\ - & \qquad\,\ - & 90.18** \\
& BPR-STNet & 2024 & 84.02$\pm$21.24* & 83.70$\pm$21.65 & 90.31$\pm$12.83 & 82.02$\pm$27.31 & 83.76$\pm$17.51* \\
& \textbf{NHGNet} & - & \textbf{93.10$\pm$7.36} & \textbf{92.61$\pm$9.06} & \textbf{91.52$\pm$11.02} & \textbf{93.01$\pm$11.15} & \textbf{91.61$\pm$8.66} \\
\midrule
\multirow{6}{*}{Dataset 2} & EEGNet & 2018 & 95.83$\pm$8.37* & 98.61$\pm$4.61 & 93.33$\pm$16.50 & 98.33$\pm$5.53 & 94.79$\pm$11.72* \\
& ICNN & 2022 & 72.36$\pm$11.81** & 78.45$\pm$15.53** & 68.33$\pm$26.75** & 76.39$\pm$22.09** & 68.90$\pm$16.26** \\
& ECNN & 2023 & 89.17$\pm$14.98** & 90.60$\pm$14.62** & 87.50$\pm$22.03** & 90.83$\pm$15.52** & 87.91$\pm$17.94** \\
& LGGNet & 2023 & 89.58$\pm$18.58** & 81.24$\pm$36.59** & 81.39$\pm$36.74** & 97.63$\pm$5.29** & 81.21$\pm$36.50** \\
& BPR-STNet & 2024 & 92.08$\pm$10.30 & 98.20$\pm$4.08 & 85.83$\pm$20.19* & 98.33$\pm$3.73 & 89.99$\pm$15.01 \\
& \textbf{NHGNet} & - & \textbf{99.25$\pm$1.90} & \textbf{99.55$\pm$1.98} & \textbf{99.00$\pm$3.27} & \textbf{99.50$\pm$2.18} & \textbf{99.23$\pm$1.97} \\
\bottomrule
\end{tabularx}
\end{small}
\end{center}
\end{table*}

\begin{table*}[t]
\caption{Inter-subject results on two datasets. Presented as mean ± standard deviation, with the highest values in bold. $p$-value indicating the improvement of NHGNet over the model: * for $p < 0.05$, ** for $p < 0.001$.}
\label{tab:inter-subject}
\begin{center}
\begin{small}
\begin{tabularx}{\textwidth}{@{}c c c X X X X X@{}} 
\toprule
Datasets & Model & Year & Accuracy(\%) & Precision(\%) & Sensitivity(\%) & Specificity(\%) & F1-score(\%) \\
\midrule
\multirow{6}{*}{\centering Dataset 1} 
& EEGNet &2018 & 80.24$\pm$8.56** & 76.48$\pm$15.06** & 81.78$\pm$15.21* & 82.91$\pm$7.10* & 76.65$\pm$9.61** \\
& ICNN &2022 & 81.39$\pm$9.71** & 84.36$\pm$12.89** & 72.83$\pm$11.27** & 90.66$\pm$7.25** & 76.94$\pm$8.45** \\
& ECNN  &2023 & 86.06$\pm$11.32** & 87.85$\pm$13.22** & 80.76$\pm$15.12** & 88.98$\pm$17.04** & 82.58$\pm$11.11** \\
& LGGNet &2023 & 84.16$\pm$10.01** & \textbf{88.63$\pm$9.40}** & 75.51$\pm$21.16** & \textbf{91.25$\pm$7.48}** & 78.69$\pm$14.06** \\
& BPR-STNet &2024 & 85.45$\pm$5.59* & 78.34$\pm$26.29* & 72.57$\pm$33.40* & 84.98$\pm$14.91 & 71.68$\pm$30.12* \\
& \textbf{NHGNet} &- & \textbf{88.15$\pm$5.82} & 84.41$\pm$9.26 & \textbf{88.60$\pm$7.03} & 85.93$\pm$12.42 & \textbf{86.01$\pm$5.98} \\
\midrule
\multirow{6}{*}{Dataset 2} 
& EEGNet &2018 & 75.67$\pm$19.18* & 77.82$\pm$19.62* & 77.17$\pm$18.37* & 74.17$\pm$27.11* & 76.44$\pm$17.56* \\
& ICNN &2022 & 69.92$\pm$9.30** & 75.89$\pm$13.14** & 63.50$\pm$19.51** & 76.33$\pm$18.05** & 66.75$\pm$11.27** \\
& ECNN &2023 & 70.50$\pm$19.49** & 76.01$\pm$20.81** & 68.33$\pm$25.79** & 72.67$\pm$29.08** & 68.55$\pm$20.96** \\
& LGGNet &2023 & 70.86$\pm$14.51** & 70.53$\pm$16.19** & 88.67$\pm$15.58** & 53.05$\pm$35.58** & 75.93$\pm$9.92** \\
& BPR-STNet &2024 & 75.00$\pm$14.45* & 76.58$\pm$14.80* & 73.33$\pm$23.78* & 76.67$\pm$17.44* & 73.17$\pm$17.34* \\
& \textbf{NHGNet} &- & \textbf{90.73$\pm$7.25} & \textbf{90.24$\pm$8.58} & \textbf{92.37$\pm$9.33} & \textbf{89.08$\pm$11.19} & \textbf{90.86$\pm$7.39} \\
\bottomrule
\end{tabularx}
\end{small}
\end{center}
\end{table*}

\subsection{Experimental Setup}
In this investigation, two publicly accessible EEG datasets for fatigue virtual driving were used to evaluate the performance of the proposed NHGNet, referred to as Dataset 1 \cite{cao2019multi} and Dataset 2 \cite{min2017driver}. Dataset 1 was preprocessed in a way consistent with the compared model ICNN \cite{cui2022eeg}, while Dataset 2 maintained the preprocessing methods as outlined in the original work. Both datasets were resampled to a frequency of 128 Hz to guarantee uniformity.
Owing to the time-consuming nature of EEG data collection, the resulting datasets are frequently small, significantly contributing to the risk of model overfitting. Traditional methods, including the application of Gaussian noise or cropping, might lead to reduced signal-to-noise ratios or alterations in the intrinsic relationships of the data. We used temporal segmentation and recombination (TSR), which included separating training data from the same category into eight equal segments and randomly recombining them while retaining the integrity of the time series. This led to a fivefold increase in data volume per category, allowing for larger sample sizes while retaining data accuracy and proportional consistency with the original samples.

This study developed NHGNet based on PyTorch, which operates on an NVIDIA GeForce RTX 2080TI GPU equipped with 11 GB of memory. We limit intra-subject and inter-subject base model training to 200 epochs. We meticulously selected batch sizes in this study to align with the distinct objectives of each experimental design. We employed a batch size of 64 for intra-subject analysis to precisely detect subtle, individual-level variations, thereby reducing the risk of overfitting. In contrast, we chose a batch size of 1024 for inter-subject train to efficiently process the larger and more heterogeneous data, thereby enhancing model stability and generalization across diverse subjects. To further prevent overfitting, we set a dropout rate of 0.5 and utilised the Adam optimizer to optimise the model. To enhance generalisation in the model, we introduced regularisation coefficients $\lambda_1$ and $\lambda_2$, assigned values of $1e-4$ and $1e-2$, respectively.

\begin{figure*}[t]
    \centering
    \includegraphics[width=\textwidth]{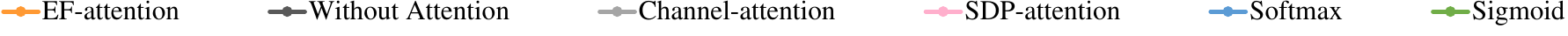}
    \subfloat[Dataset 1: Intra-Object]{\includegraphics[width=0.24 \textwidth]{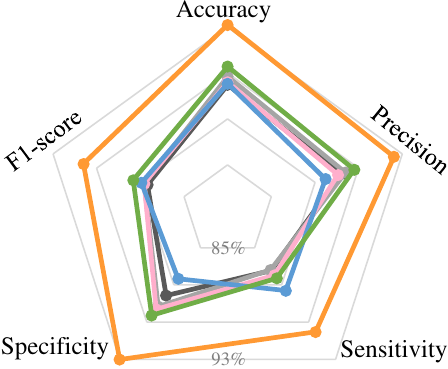}\label{fig:chenq2_a}}\hfill
    \subfloat[Dataset 2: intra-object]{\includegraphics[width=0.247\textwidth]{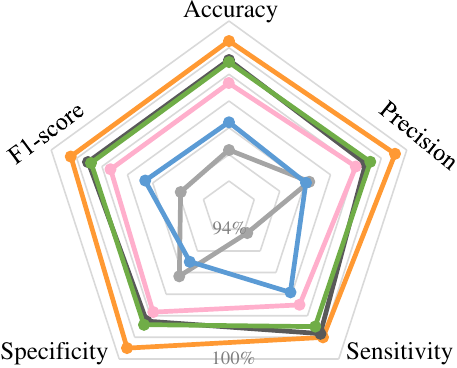}\label{fig:chenq2_b}}\hfill
    \subfloat[Dataset 1: inter-object]{\includegraphics[width=0.247\textwidth]{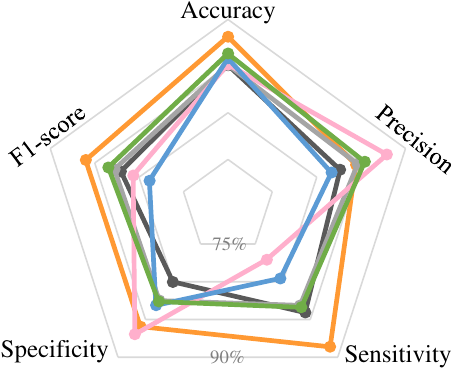}\label{fig:chenq2_c}}\hfill
    \subfloat[Dataset 2: inter-object]{\includegraphics[width=0.247\textwidth]{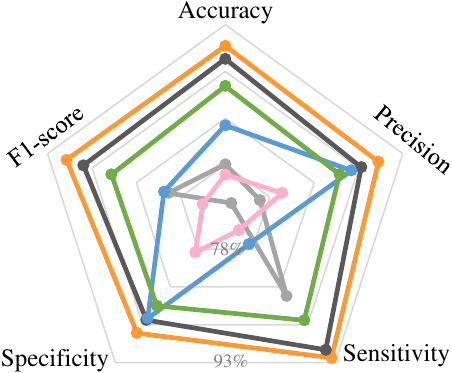}\label{fig:chenq2_d}}
    \caption{The subtitle provides an elucidation of the data sources and the experimental approach. The curves of different colors in each figure represent various settings of attention mechanisms or activation functions.}
    \label{fig:attention}
\end{figure*}

\subsection{Experimental Evaluation}
\subsubsection{Model Comparison}
Our study compared the proposed NHGNet against state-of-the-art deep learning methods including EEGNet \cite{lawhern2018eegnet}, ICNN \cite{cui2022eeg}, ECNN \cite{li2023decomposition}, LGGNet \cite{ding2023lggnet}, and BPR-STNet \cite{lin2024eeg} on Dataset 1 and Dataset 2. The source articles for these models include detailed descriptions of their network designs and implementation methodologies. We conducted experiments on two distinct datasets, both intra-subject and inter-subject, enabling us to perform a comparative analysis of model performance. The evaluation criteria encompassed metrics such as accuracy, precision, sensitivity, specificity, and the f1-score.

All baseline models were trained using optimal parameters and training procedures predetermined by the original text to maintain consistency and fairness in the experiments. In the inter-subject studies, following the training of the basic models, all models applied the same transfer learning technique as NHGNet to ensure equitable comparisons.

In the comparative analysis conducted on Table \ref{tab:intra-subject}, NHGNet demonstrated significant competitive advantages in the performance of fatigue driving detection across two datasets in intra-object experiments. Specifically, the NHGNet exhibited exceptional stability and reliability, evidenced by its highest scores across various key performance indicators. In dataset 1, NHGNet achieves an accuracy of 93.10\%, surpassing the suboptimal model, LGGNet, by 2.34\%, and outperforms all other models in all other important performance parameters. Dataset 2 demonstrates superior performance by NHGNet, with an accuracy and f1-score of 99.25\% and 99.23\%, respectively. This implies that the NHGNet identified nearly all instances of fatigue driving. The accuracy of the model EEGNet is 3.42\% higher compared to the suboptimal model, and the f1-score is 4.44\% higher.

The cross-object results as shown in Table \ref{tab:inter-subject} indicate that for Dataset 1, NHGNet achieved an accuracy of 88.15\%$\pm$5.82, significantly surpassing other models. Although our model did not achieve the highest values in precision and specificity, it still demonstrated competitive performance levels. In contrast, for the fatigue driving detection in datasets with imbalanced labels, metrics such as accuracy, sensitivity, and F1 score are crucial as they reflect the model's high level of recognition for the fatigue label of interest. For Dataset 2, our model's accuracy reached 90.73\%, exceeding the next best model, EEGNet, by 15.06\%. Moreover, it outperformed the second-highest models in terms of precision, sensitivity, specificity, and F1 score by 12.42\%, 3.7\%, 12.42\%, and 14.43\%, respectively, with statistical significance.
\begin{figure}[htbp]
    \centering
    \subfloat[Dataset 1]{\includegraphics[width=0.8\linewidth]{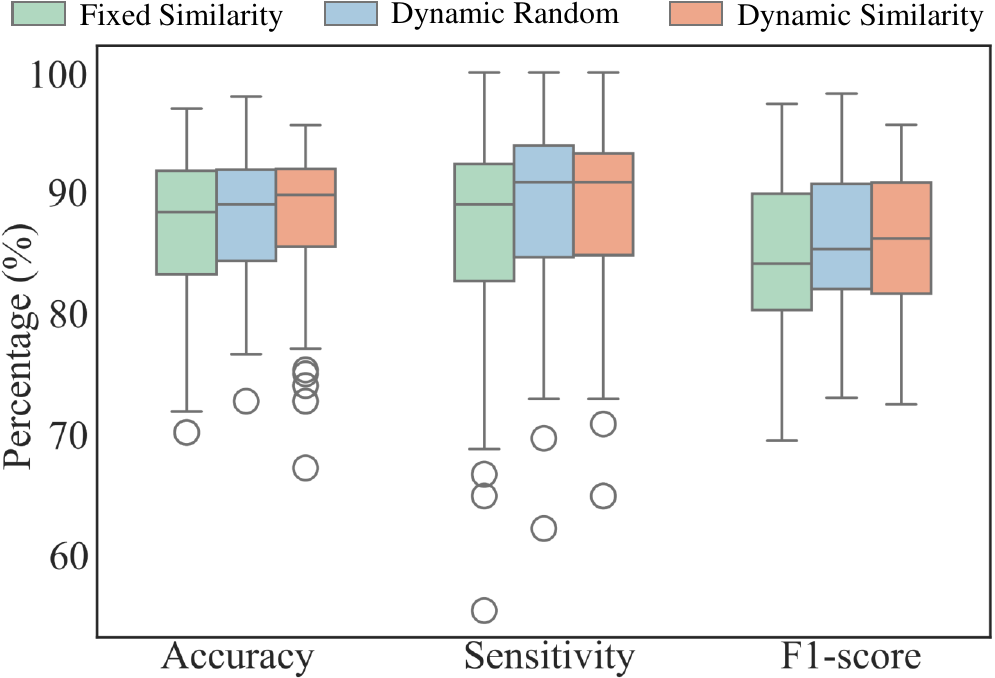}\label{fig:AdjacencyMatrix_Dataset1}} \\
    \subfloat[Dataset 2]{\includegraphics[width=0.8\linewidth]{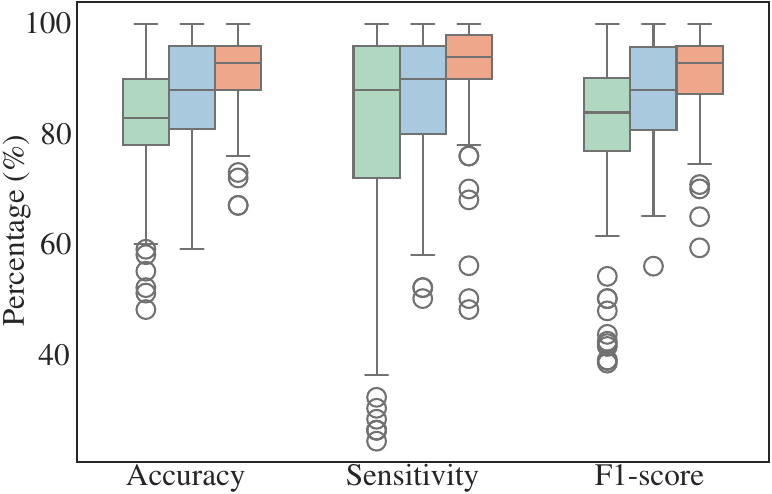}\label{fig:AdjacencyMatrix_Dataset2}}
    \caption{Inter-subject evaluation of adjacency matrix variants on two datasets. Comparative performance of dynamic similarity, dynamic random, and fixed similarity adjacency matrices in terms of accuracy, sensitivity, and f1-score.}
    \label{fig:AdjacencyMatrixVariants}
\end{figure}

\subsubsection{Attention Experiment}
In assessing the efficacy of the EF-attention mechanism, this study employed a controlled variable method, conducting comparative performance analyses on two distinct datasets. The experiments were organized as follows: initially, the attention structures within the network model were removed; subsequently, two classic attention mechanisms, namely Channel-attention \cite{gao2023sft} and scaled dot-product attention (SDP-attention) \cite{song2022eeg,fu2024spatiotemporal}, were compared; finally, to ascertain the impact of activation functions on model performance, the activation function within EF-attention was altered from $\text{tanh}$ to $\text{softmax}$ and $\text{Sigmoid}$. The experimental results are illustrated in the Fig. \ref{fig:attention}.

As observed in Fig. \ref{fig:attention}, models utilizing EF-attention demonstrated a clear advantage, especially in Fig. \ref{fig:chenq2_a}, Fig. \ref{fig:chenq2_b}, and Fig. \ref{fig:chenq2_d}. Although in Fig. \ref{fig:chenq2_c}, EF-attention did not achieve optimal precision, its overall performance still surpassed other models, as indicated by the largest area of the radar chart generated. From the perspective of the F1-score, a comprehensive evaluation metric, EF-attention exceeded the next best model by $2.28\%$ and $0.68\%$ intra-subject in the two datasets, respectively, and outperformed by $2.55\%$ and $1.89\%$ in inter-subject models.

\begin{figure*}[t]
    \centering
    \subfloat[]{\includegraphics[width=0.23\linewidth]{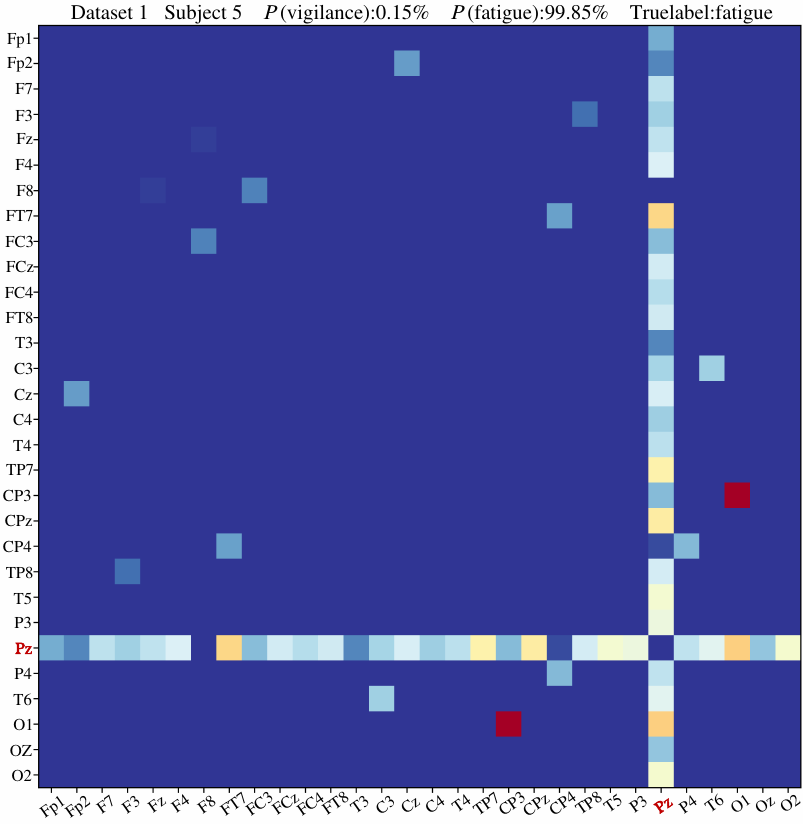}\label{fig:Dataset1_Sub5_Fatigue}}
    \hfill
    \subfloat[]{\includegraphics[width=0.23\linewidth]{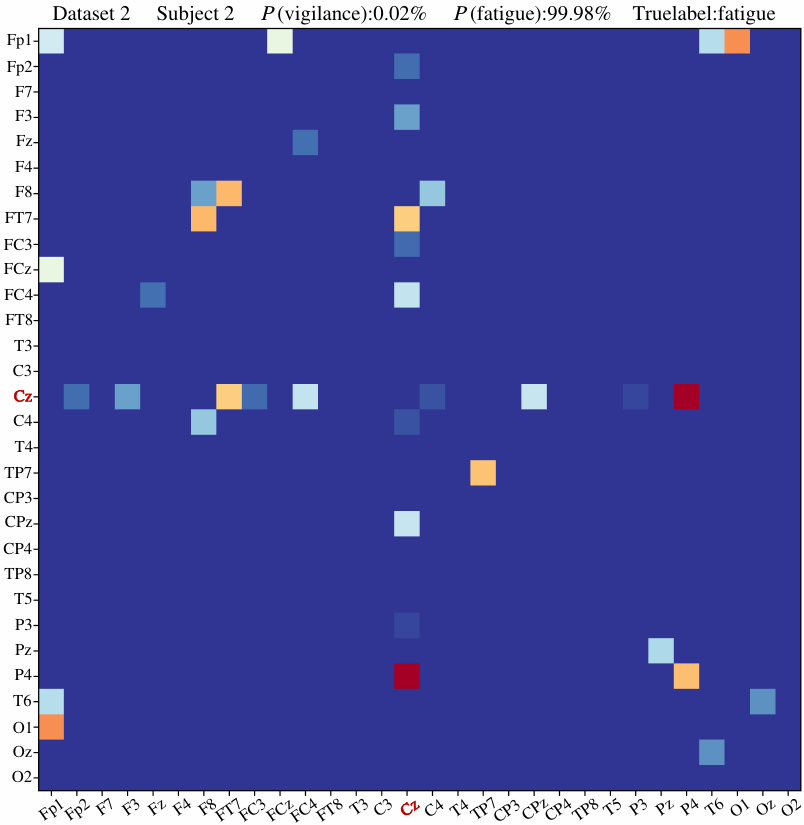}\label{fig:Dataset2_Sub2_Fatigue}}
    \hfill
    \subfloat[]{\includegraphics[width=0.23\linewidth]{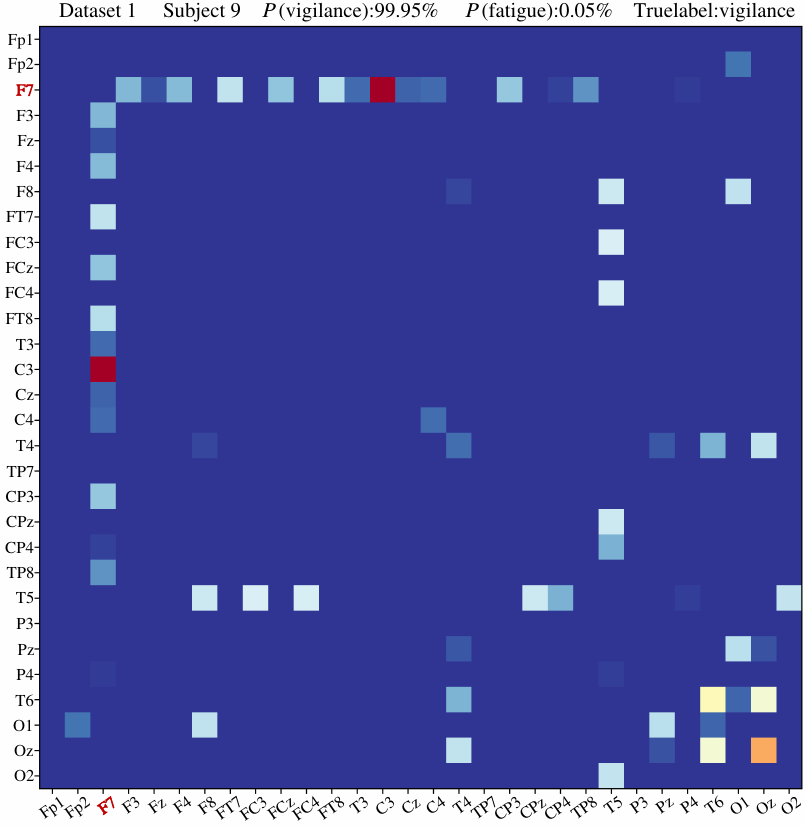}\label{fig:Dataset1_Sub9_Vigilance}}
    \hfill
    \subfloat[]{\includegraphics[width=0.23\linewidth]{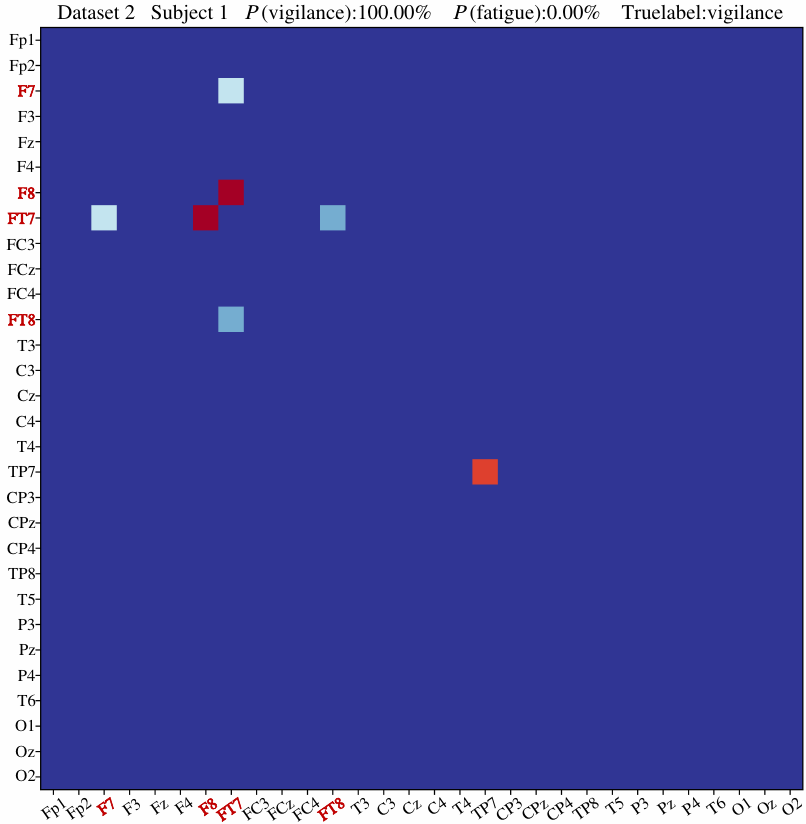}\label{fig:Dataset2_Sub1_Vigilance}}
    \hfill \includegraphics[height=0.17\textheight]{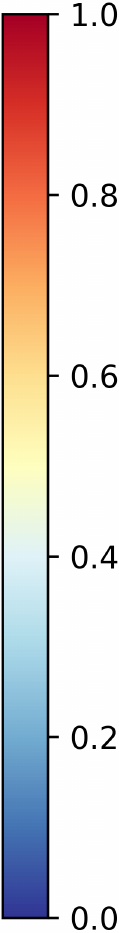} 
    \caption{Visualization of EEG channel correlations across different datasets and subjects. Each subplot is marked with the data source, predicted probabilities for vigilance and fatigue stages from NHGNet, and the true labels of the samples.}
    \label{fig:AdjacencyMatrix}
\end{figure*}

\begin{figure}[t]
    \centering
    \subfloat[]{\includegraphics[width=0.46\columnwidth]{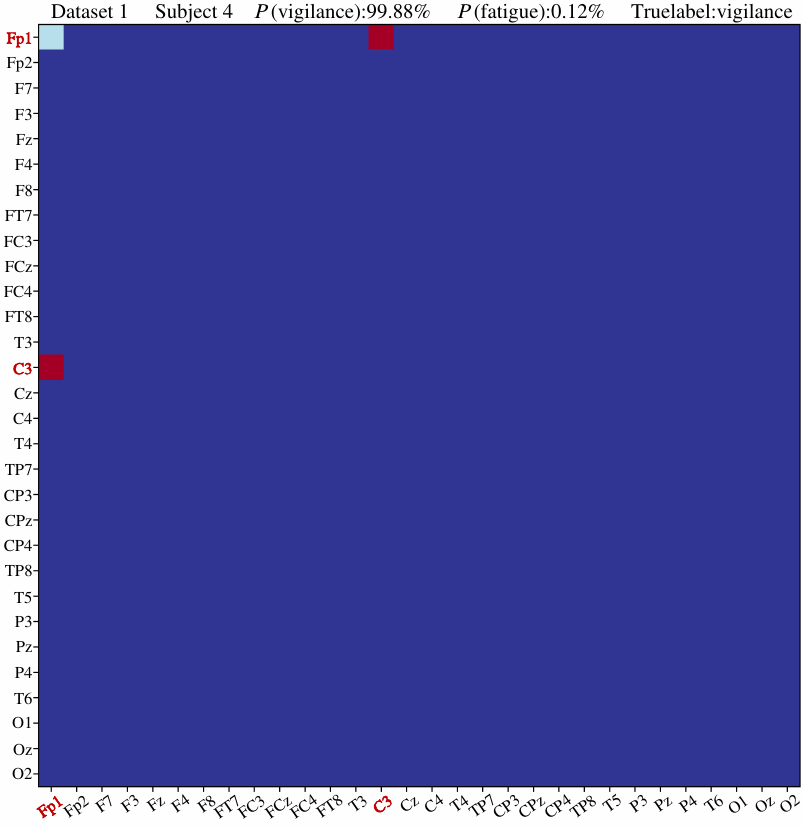}\label{fig:Dataset1_Sub4_Vigilance_C3}}
    \hfill
    \subfloat[]{\includegraphics[width=0.46\columnwidth]{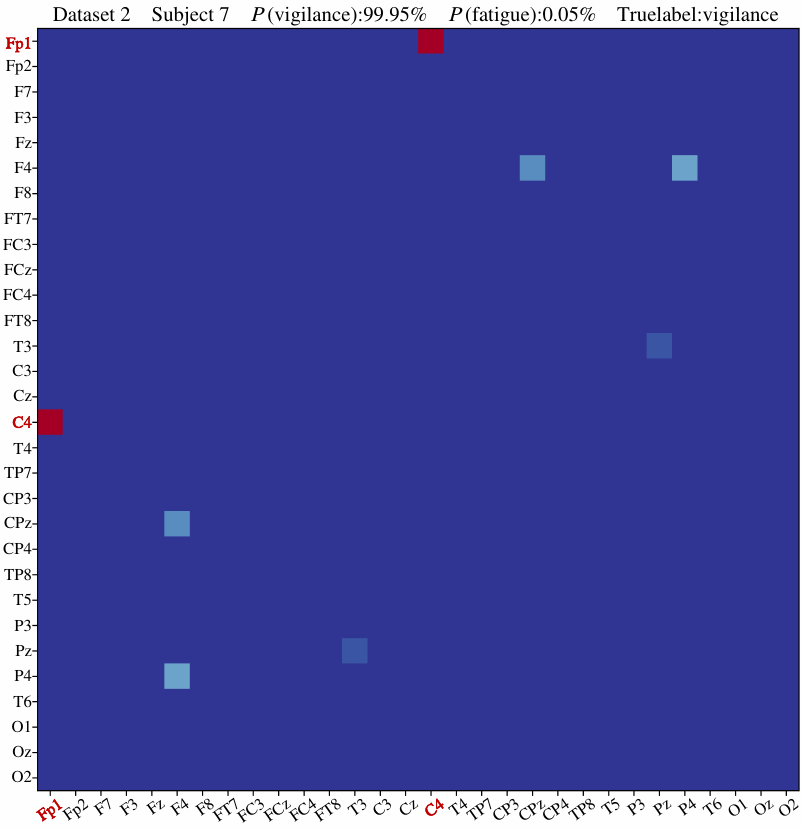}\label{fig:Dataset2_Sub7_Vigilance_C4}}
    \hfill    {\includegraphics[height=0.17\textheight]{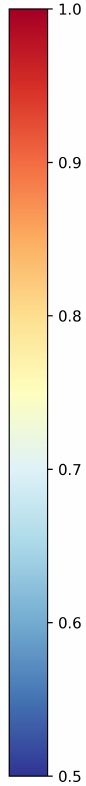}\label{fig:Dataset_Extra_Image}} 
    \caption{Visualization of the relationship between EEG channels in vigilance samples from two datasets. Each subplot, marked at the top, shows the data source, prediction probabilities for vigilance and fatigue states from NHGNet, and the true labels.}
    \label{fig:AdjacencyMatrix_C3C4}
\end{figure}

\section{Discussion}

\subsection{Adjacency Matrix Variants}
This section compares several AM configurations, especially dynamic similarity, dynamic randomness, and fixed similarity, between two datasets in an inter-subject experiment. The objective is to examine the impact of trainability and initialization methods on the model's performance, as detailed in Fig. \ref{fig:AdjacencyMatrixVariants}. In Dataset 1, making the AM trainable consistently resulted in better performance, demonstrating the efficacy of dynamically modifying the AM using backpropagation to better capture the inter-channel correlations in the data. For Dataset 2, the similarity-initialized dynamic AM performed better, demonstrating that similarity-based initialization aids the model in leveraging the structural properties of the data during the initial training phase.

However, when the adjacency matrices were made to not be trainable, both datasets showed a considerable performance reduction. This conclusion emphasizes the importance of trainability in optimizing models to capture complicated data aspects. A fixed AM, even with proper initialization, may lack flexibility due to its inability to respond to new information throughout the training process.

In summary, we can confirm the effectiveness of making the AM trainable. The approach of initializing with similarity proved successful in our studies, providing the model with a strong starting point based on prior knowledge of the data structure. Furthermore, for instances where there is no clear previous understanding of the data structure at the initialization step, the strategy of setting the matrix as trainable after random initialization has demonstrated its viability. This strategy allows the model to learn and adjust independently during the training phase.

\subsection{Neural Area Interactions}
The previous section confirmed the effectiveness of setting the AM as trainable. Next, we will explore the connections between EEG channels under different cognitive states by visualizing the trained AM. Fig. \ref{fig:AdjacencyMatrix} visualizes adjacency matrices for reliably identified samples from two datasets and subjects. For fatigue state samples Fig. \ref{fig:Dataset1_Sub5_Fatigue} and Fig. \ref{fig:Dataset2_Sub2_Fatigue}, there is a discernible and significant correlation between the Pz and Cz channels located in the central brain region and most other channels. This suggests that the Pz and Cz channels play a crucial role in fatigue-state discrimination.

In the vigilance sample Fig. \ref{fig:Dataset1_Sub9_Vigilance}, the F7 channel stands out from the rest of the others. Moreover, the results for sample Fig. \ref{fig:Dataset2_Sub1_Vigilance} highlight the tight linkages between the frontal and temporal lobe channels, specifically between FT7 and F7, FT7 and F8, and within the temporal lobe between FT8 and FT7. This marked interaction between the frontal and temporal lobes accentuates their collaborative function in maintaining vigilance.

\begin{figure*}[t]
    \centering
    \begin{minipage}{\linewidth} 
        \centering 
        \subfloat[Dataset 1 Subject 2]{\includegraphics[width=0.23\linewidth]{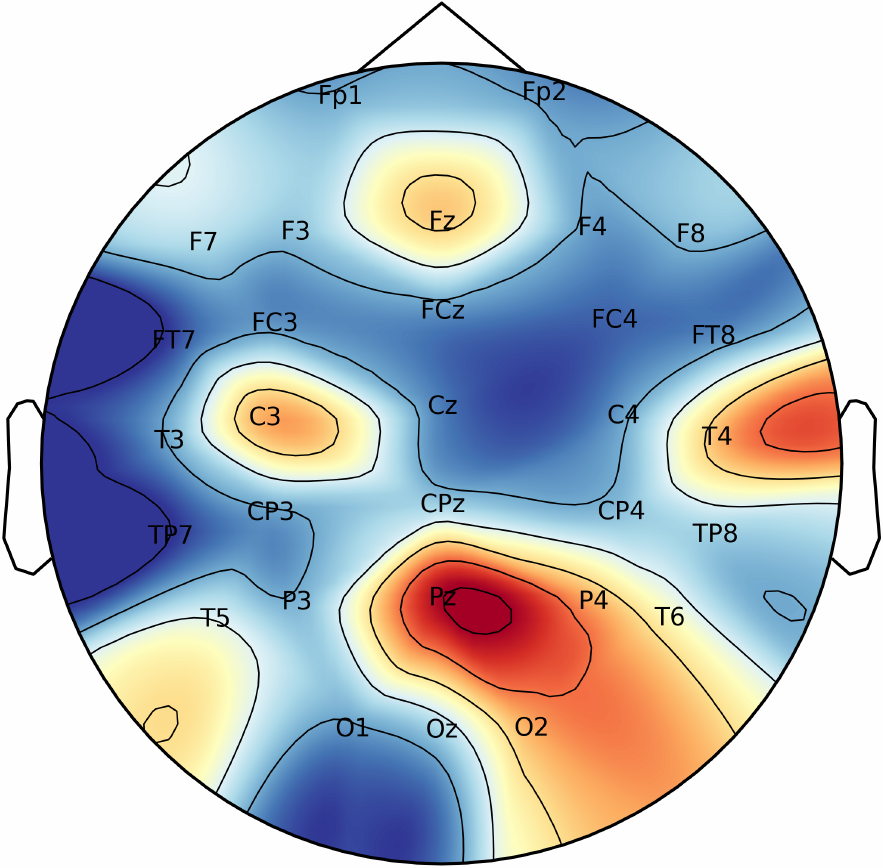}\label{fig:Dataset1_Sub_2_Fatigue}}
        \hfill 
        \subfloat[Dataset 1 Subject 11]{\includegraphics[width=0.23\linewidth]{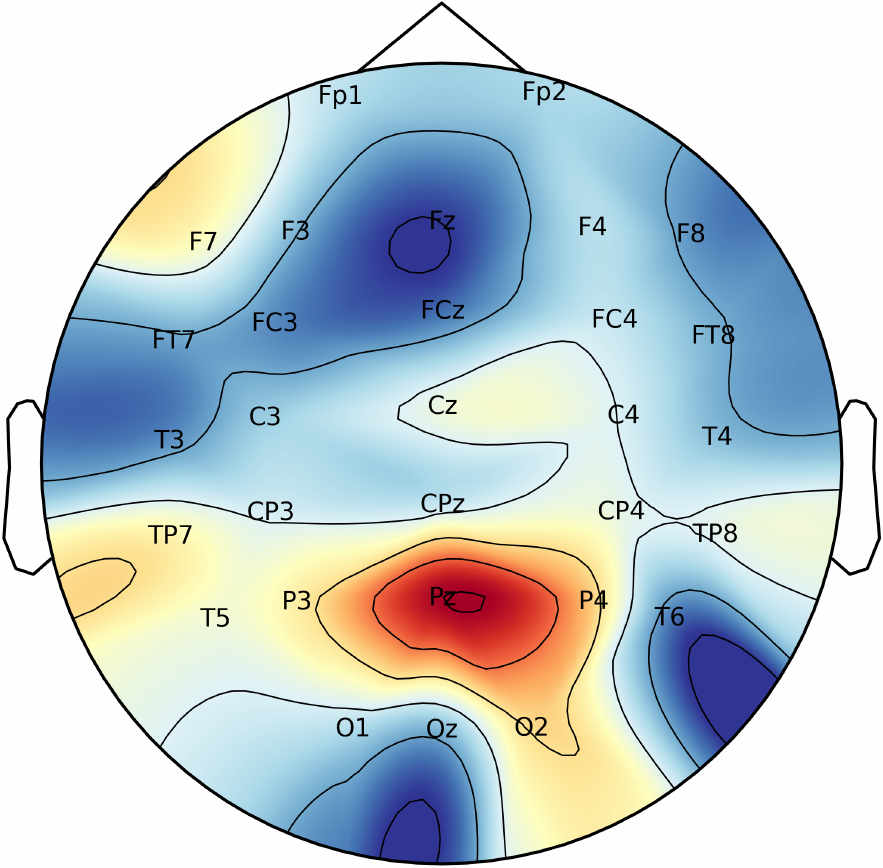}\label{fig:Dataset1_Sub_11_Fatigue}}
        \hfill 
        \subfloat[Dataset 2 Subject 6]{\includegraphics[width=0.23\linewidth]{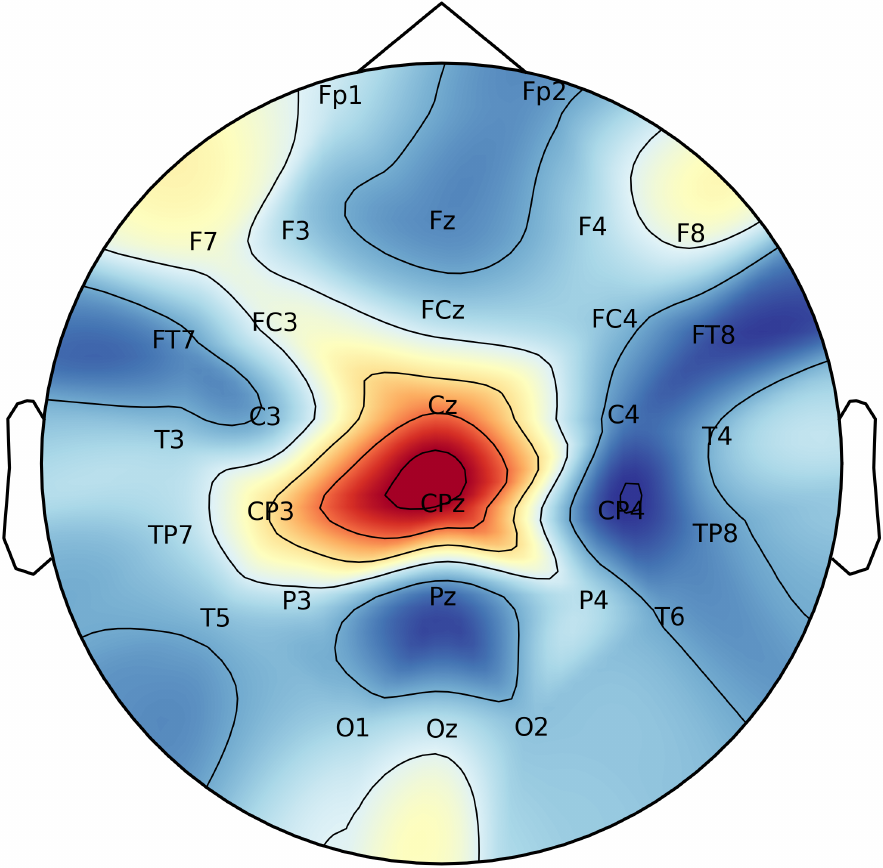}\label{fig:Dataset2_Sub_6_Fatigue}}
        \hfill 
        \subfloat[Dataset 2 Subject 9]{\includegraphics[width=0.23\linewidth]{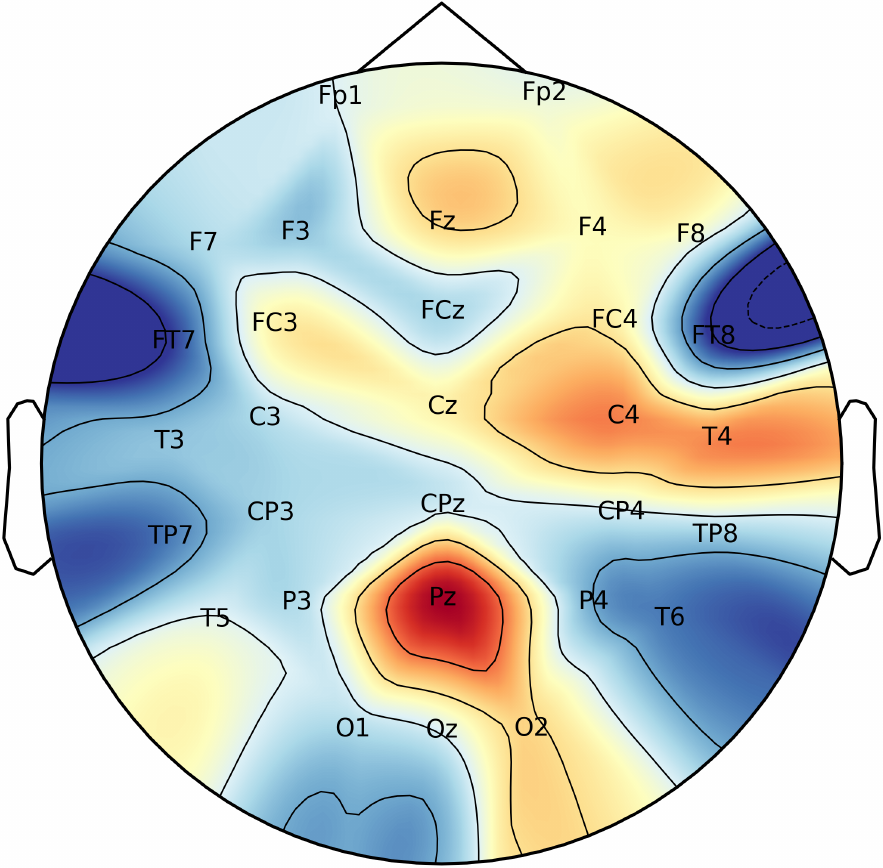}\label{fig:Dataset2_Sub_9_Fatigue}}
        \hfill \includegraphics[height=0.175\textheight]{chenq4_e.pdf} 
    \end{minipage}
    \caption{Visualization analysis of spatial feature extractor outputs from NHGNet in fatigue EEG sample classification. The output features were normalized to produce the topographic heat maps shown above, with fatigue prediction probabilities of 99.77\%, 98.81\%, 99.92\%, and 85.93\%, respectively.}
    \label{fig:ChannelWeight_Fatigue}
\end{figure*}

The visualization analysis of vigilance-related EEG signals under simulated driving scenarios (as Fig. \ref{fig:AdjacencyMatrix_C3C4} illustrated) revealed a notable correlation between the Fp1 channel in the frontal lobe and the central region channels C3 and C4, which are associated with motor control \cite{gwon2021alpha,shahlaei2023quantification}. This finding is consistent with past studies Fig. \ref{fig:ChannelWeight_Fatigue}, which highlight the vital role of channels C3 and C4 in completing the hand and upper limb. In driving circumstances that demand rapid responses, a vigilance state typically accompanies the frequent rotation of the steering wheel, requiring the full attention and coordination of the driver.

\begin{figure*}[t]
    \centering
    \begin{minipage}{\linewidth} 
        \centering 
        \subfloat[Dataset 1 Subject 5]{\includegraphics[width=0.23\linewidth]{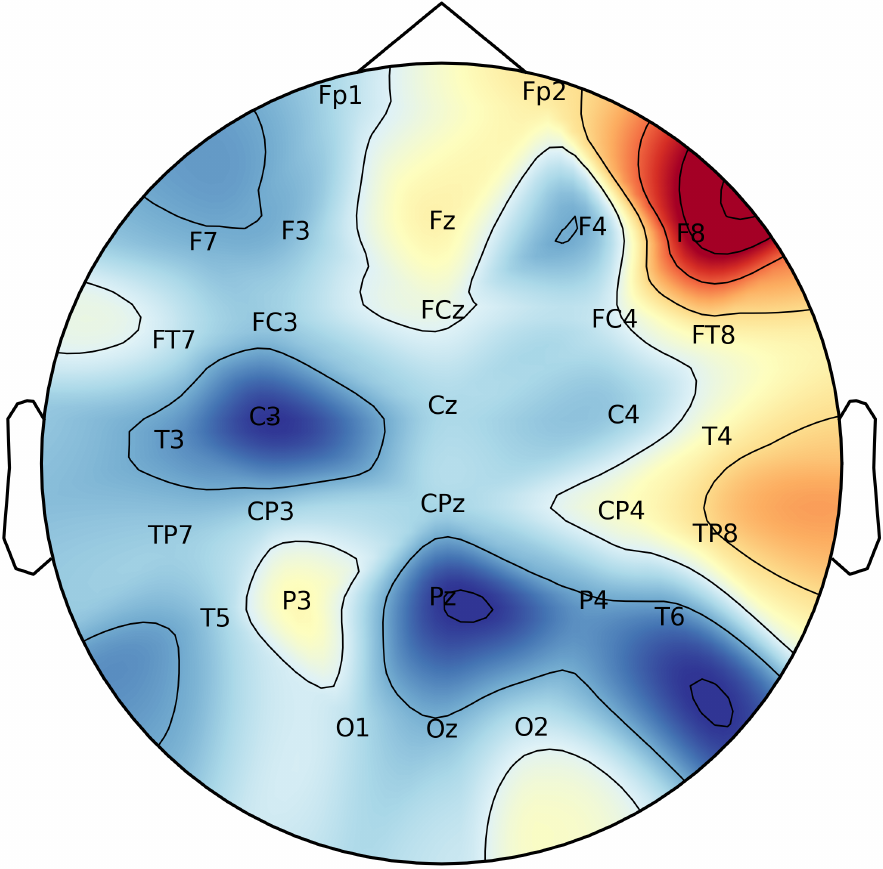}\label{fig:Dataset1_Sub_5_Vigilance}}
        \hfill 
        \subfloat[Dataset 1 Subject 9]{\includegraphics[width=0.23\linewidth]{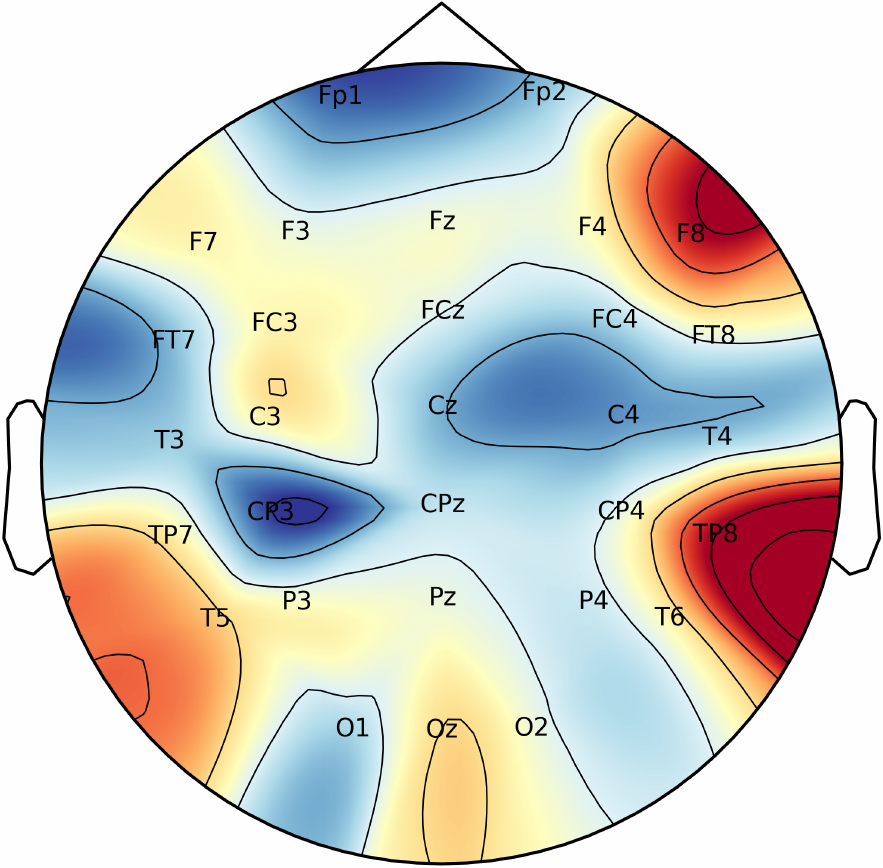}\label{fig:Dataset1_Sub_9_Vigilance}}
        \hfill 
        \subfloat[Dataset 2 Subject 1]{\includegraphics[width=0.23\linewidth]{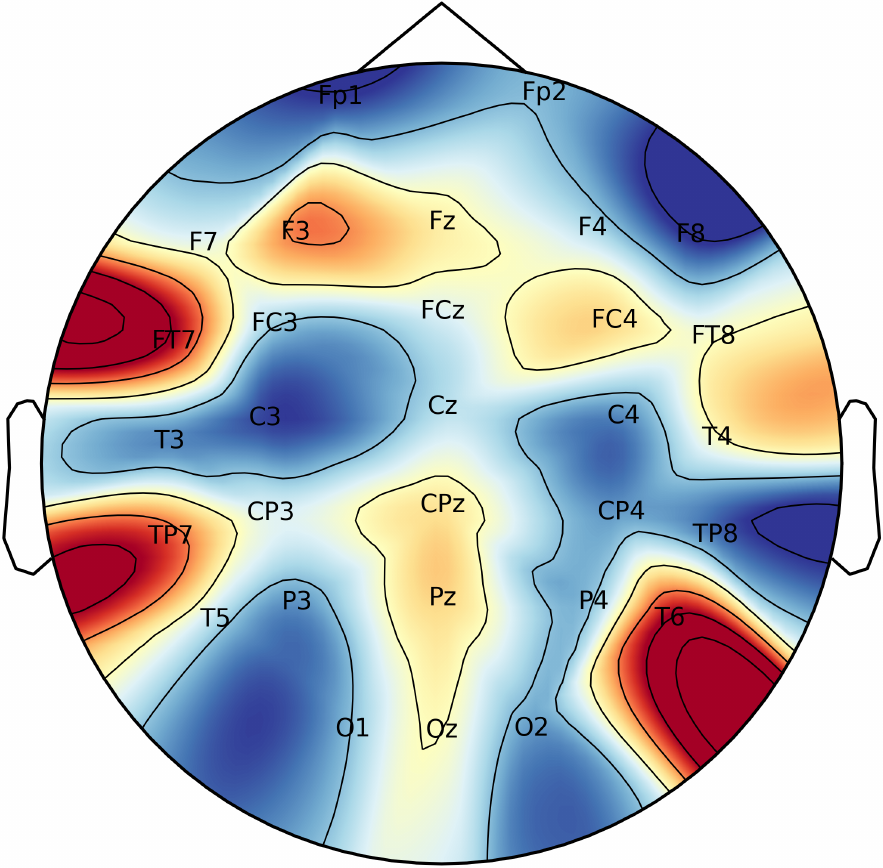}\label{fig:Dataset2_Sub_1_Vigilance}}
        \hfill 
        \subfloat[Dataset 2 Subject 2]{\includegraphics[width=0.23\linewidth]{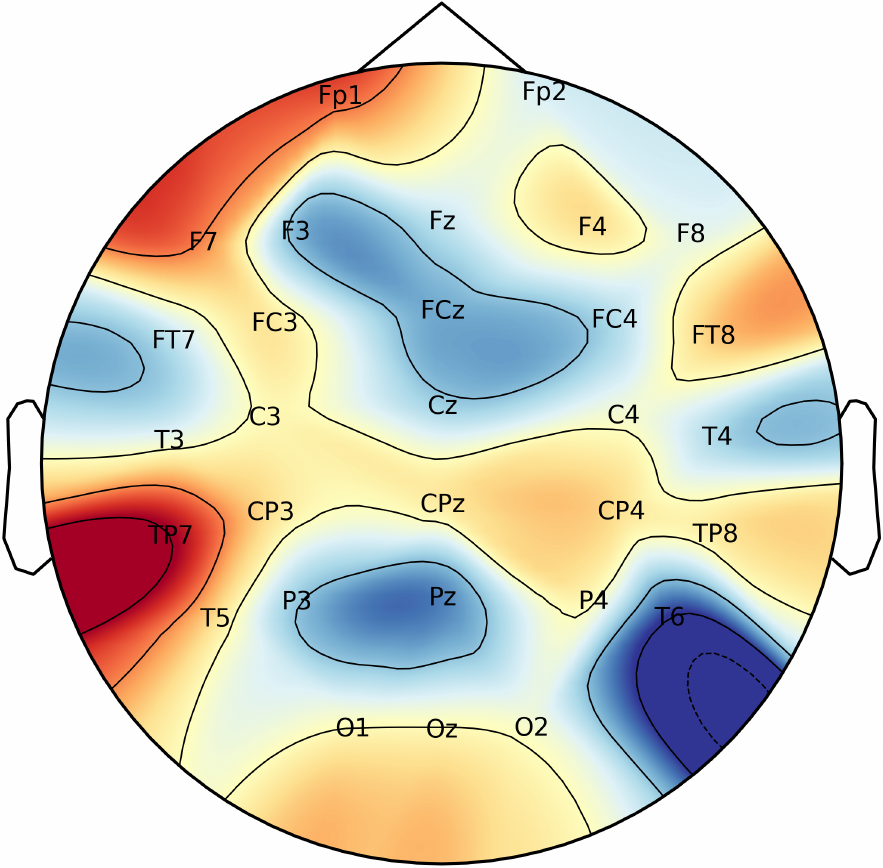}\label{fig:Dataset2_Sub_2_Vigilance}}
        \hfill \includegraphics[height=0.175\textheight]{chenq4_e.pdf} 
    \end{minipage}
    \caption{Visual analysis of spatial feature extraction outputs by NHGNet for vigilance state EEG sample classification. The feature output, after normalization, has been rendered into the topographic heat maps presented, with the predictive probabilities for the vigilance state reaching 100.0\%, 100.00\%, 99.99\%, and 99.99\%, respectively.}
    \label{fig:ChannelWeight_Vigilance}
\end{figure*}

\subsection{Critical Brain Regio}
In this section, we explore how NHGNet differentiates between states of fatigue by analyzing EEG signals. We showcase representative samples exhibiting common characteristics from various subjects across different datasets, which are accurately classified with high confidence as indicative of drowsy and alert states, as detailed in Fig. \ref{fig:ChannelWeight_Fatigue} and Fig. \ref{fig:ChannelWeight_Vigilance}, respectively. Fig. \ref{fig:ChannelWeight_Fatigue} presents a topological heatmap that visualises the spatial features extracted by the model from EEG samples in a state of fatigue, highlighting significant activity intensity in the Pz and CPz channels. This underscores their importance in identifying the state of fatigue. This finding aligns with existing literature \cite{cui2022eeg,li2023decomposition}, indicating that enhanced activity in these specific channels is a key physiological marker of fatigue. 

In Fig. \ref{fig:ChannelWeight_Vigilance}, the display of spatial feature extraction from vigilance EEG recordings reveals significant activity patterns in the frontal and temporal lobes during vigilance states. The frontal lobe uses channels like Fp1, F7, F8, and FT7 to maintain attention, working memory, and other executive tasks \cite{wang2020multiple,wang2021eeg}. The temporal lobe is active in processing environmental stimuli, particularly in quickly responding to external information, which is directly related to vigilance.

\section{Conclusion}
In this study, we focused on the spatial features of non-Euclidean electroencephalogram (EEG) data to increase fatigue detection accuracy. We introduced the node-holistic graph convolution network (NHGNet), which uses a dynamic graph convolutional network to convey spatial information inside each brain functional region while also capturing the complex relationships between different brain areas. Besides, exact fit attention is used to assign appropriate weights to various EEG channels. On two publicly available datasets, NHGNet outperforms numerous state-of-the-art algorithms in terms of accuracy and f1-score. Furthermore, visualization techniques reveal that channels Pz, CPz, and Cz are key for recognizing fatigue. In vigilance, frontal and temporal lobe EEG channels become more prominent, and Fp1 is correlated with movement control C3 and C4 channels. In summary, we provide key insights into the function of EEG channels in cognitive state detection, and future research focuses on applying neuropsychological knowledge to deep learning frameworks.

\bibliographystyle{IEEEtran}
\bibliography{NHGNet}
\end{document}